\documentclass[a4paper,11pt]{article}

\pdfoutput=1
\usepackage{amsfonts}
\usepackage{tabularx}
\usepackage{booktabs}
\usepackage{amssymb}
\usepackage[backref=page, hidelinks]{hyperref}
\usepackage{graphicx}
\usepackage{multirow}
\usepackage{multicol}
\usepackage{color}
 \usepackage{graphicx}
\usepackage{url}
\usepackage[affil-it]{authblk}
\usepackage{amsmath}
\usepackage{wrapfig}
\usepackage{color, colortbl}
\usepackage{ragged2e}
\usepackage{breakcites} %Alan added this to fix citations not wrapping to a new line in the text thus pushing outside margins

\usepackage[T1]{fontenc}
\usepackage{times}
\usepackage{todonotes}

\begin{document}

\title{Dynamic Channel Selection in Self-Supervised Learning}

\author{Tarun Krishna\textsuperscript{\ref{eq}}, ~Ayush K. Rai\textsuperscript{\ref{eq}}, ~Yasser A. D. Djilali, Alan F. Smeaton,\\ Kevin  McGuinness and Noel E. O'Connor}
\affil{Insight Centre for Data Analytics, Dublin City University, Dublin, Ireland}
\date{}
\maketitle
\def\thefootnote{*}\footnotetext{\label{eq}Equal contribution.}
\thispagestyle{empty}
\definecolor{LightCyan}{rgb}{0.88,1,1}

\begin{abstract}
Whilst computer vision models built using self-supervised approaches are now commonplace, some important questions remain. Do self-supervised models learn highly redundant channel features? What if a self-supervised network could dynamically select the important channels and get rid of the unnecessary ones? Currently, convnets pre-trained with self-supervision have obtained comparable performance on downstream tasks in comparison to their supervised counterparts in computer vision. However, there are drawbacks to self-supervised models including their large numbers of parameters, computationally expensive training strategies and a clear need for faster inference on downstream tasks. In this work, our goal is to address the latter by studying how a standard channel selection method developed for supervised learning can be applied to networks trained with self-supervision. We validate our findings on a range of target budgets $t_{d}$ for channel computation on image classification task across different datasets, specifically CIFAR-10, CIFAR-100, and ImageNet-100, obtaining comparable performance to that of the original network when selecting all channels but at a significant reduction in computation reported in terms of FLOPs. 

% We further evaluate whether selecting fewer channels during pre-training using channel gating in the whole network based on a target budget $t_d$ obtains comparable performance to that of the original network when selecting all channels.

%In experiments, we pretrain Simsiam model \cite{chen2020simple} with a ResNet18 backbone on CIFAR10, CIFAR100 and ImageNet100 datasets and perform downstream evaluation on weighted k-nearest neighbour task. Our results show that.........% performance close to td = 1.
%We also observe that using a channel selection mechanism can reduce inference time on downstream task without excessively increasing the pre-training time. Our extensive experiments show that our approach can save x\% of computation reported in terms of hardware independent theoretical metric of FLOPs. 

    %Noel's comment
    %Results are reported in terms of theoretical metric FLOPs and show .....
\end{abstract}

%%%%%%%%%%%%%%%%%%%%%%
\section{Introduction}
Self-supervised pre-training of convolutional neural networks, as mentioned in \cite{chen2020simple}, has almost matched the performance of supervised pre-training on the ImageNet \cite{deng2009imagenet} image classification task, but at a cost of a huge number of parameters and inefficient training and inference methods. In the supervised learning setting, as described in \cite{veit2018convolutional}, it is accepted that networks with dynamic data dependent (conditional) channel computation architectures  during inference can lead to enhanced representation power, adaptivity, interpretability and can greatly reduce  computation cost and memory resources without compromising on the accuracy by a significant margin. This motivates us to investigate the behaviour of neural networks with a channel selection mechanism trained under self-supervision. We hypothesise that self-supervised models are an ideal candidate for such dynamic network structures as they capture highly redundant channel features during pre-training. In addition, there is also a great need to explore more efficient inference methods on downstream tasks for SSL.

In order to establish the trade off between computation and performance, there are two well established research directions when it comes to introducing channel sparsity using dynamic structure in neural networks: channel pruning and channel conditional computation. Dynamic channel pruning, as reported in \cite{gao2018dynamic}, estimates channel saliency measures and allows a network to learn and prioritise certain channels and ignore the irrelevant ones given a fixed target density.  Models based on pruning usually learn sparsity through a three stage pipeline i.e,~ pretrain-prune-finetune while in other works like \cite{tiwari2021chipnet} the pruning stage itself consists of two steps, namely soft pruning and hard pruning. Conditional channel computation as proposed in \cite{herrmann2020channel} learns to compute only a subset of channels in every layer for the given input and hence benefits inference time efficiency and provides an insight into dataset specific network behaviour. Both channel pruning and conditional channel computation are categorical decisions that cannot be optimised by gradient descent methods; however, using the Gumbel-Softmax trick from \cite{jang2016categorical} provides a way to overcome this challenge. Adafuse \cite{meng2020adafuse} proposed an adaptive temporal fusion network that learns a decision policy to dynamically fuse channels from current and history feature maps (i.e. dynamically deciding which channels to keep, reuse or skip per layer and per instance) for action recognition. Notwithstanding these works, the use of dynamic networks for channel selection has to date been mostly limited to supervised learning settings only.

To the best of our knowledge, there is no study on the impact of conditional channel selection on SSL. The work described in \cite{caron2020pruning} studies the effect of standard pruning techniques developed for supervised learning on a network trained with self-supervision. In particular they use an iterative magnitude based pruning technique described in \cite{han2015learning}, which compresses the network by alternatively minimising a training objective and pruning the network parameters with smallest magnitude. The weights of the resulting sub-network are reset based on a weight initialisation scheme: the lottery winning ticket \cite{frankle2018lottery,frankle2020linear}. We adopt a similar strategy but focus on exploring the application of standard conditional channel selection methods, as proposed in \cite{Li2021DynamicDG}, to self-supervised models during the pre-training stage and do not include any re-training. Our contributions can be summarised as follows:

\begin{enumerate}
    \item \textit{Do self-supervised models learn redundant channel features}? Through our exhaustive evaluation we demonstrate that  the SSL model (SimSiam) does indeed learn redundant channel features. 
    \item We show in Table~\ref{tab2} that exploiting this redundancy leads to a drop in computational complexity (FLOPs), reducing inference time without excessively increasing training time, as we learn from scratch and on-the-fly, unlike competing approaches \cite{caron2020pruning} that involve re-training.
    \item We demonstrate that this channel selection mechanism preserves the feature quality when evaluated on the task of image classification and gives comparable performance when compared with a vanilla (no channel selection) SSL approach. 
\end{enumerate}   %i.e. training from scratch. % Really Important to mention this #TODO                               

% Will write this if there is space left
%The rest of the paper is organized as follows : 

\section{Related Works}

\subsection{Self-Supervised Representation Learning}
SSL has recently matched the performance of supervised learning on several computer vision benchmarks \cite{chen2020simple, Djilali_2021_ICCV, krishna2021evaluating, bachman2019learning, grill2020bootstrap}. Below we summarize the recent works in the direction of self-supervised learning. 

\textbf{Contrastive Learning.} Contrastive learning refers to learning by comparison \cite{oord2018representation, chen2020simple} where the final objective is based on some variation of Noise Contrastive Estimation (NCE). The main intuition is to bring similar instances closer in the embedding space while contrasting them with other negative samples to avoid feature  collapse. These methods are usually trained in a Siamese setting with shared weights using a large batch size or memory bank \cite{chen2020simple, oord2018representation, wu2018unsupervised, misra2020self}. 
% They perform well but they require  large amount of negative samples.
    
    \textbf{Clustering Methods.} One category of self-supervised methods for representation learning is based on clustering \cite{caron2018deep, asano2019self, caron2020unsupervised}, which alternates between clustering the representations and learning to predict the cluster assignment. 
    % SwAV  \cite{caron2020unsupervised} builts upon \cite{asano2019self} i.e, it combines clustering with siamese networks and performs online clustering under a balanced partition constraint for each batch using the Sinkhorn-Knopp transform \cite{cuturi2013sinkhorn}.
    These clustering method are also based on contrastive approaches but at cluster level, which also makes the training computationally expensive.
    % \cite{gidaris2020learning} (and its variants OBoW \cite{gidaris2020online})
    
    \textbf{Distillation Methods.} Recent approaches like BYOL \cite{grill2020bootstrap} and SimSiam \cite{chen2021exploring},  need no negative samples, yet they learn useful representations and perform on-par with other SSL methods. They learn in a student-teacher setting and consequently avoid feature collapse. However, why and how they avoid collapse is still unclear and an open research area. 
    
    \textbf{Information Maximization.} A more principled way to avoid feature collapse is to capture information bottlenecks as in Barlow Twins \cite{zbontar2021barlow} and 
    %W-MSE \cite{ermolov2021whitening}
    VICReg \cite{bardes2021vicreg}.
    % These methods don't require any trick and are learned in symmetric siamese setting without any negative samples.

It is unclear how many channel redundant features are  learned by these self-supervised approaches. 
In this work we aim to study this redundancy by exploiting a dynamic channel selection mechanism from the literature.

\subsection{Dynamic Channel Computation}
\textbf{Channel Pruning.} 
% The idea of introducing structured and unstructured sparsity in deep learning through channel pruning to reduce memory, bandwidth and computational cost is fast growing research field today.
Channel pruning estimates channel saliency measures and eliminates all input and output connections from unimportant channels. The approach reported in \cite{wen2016learning} added group Lasso on channel weights to the model's training loss function resulting in a reduction of the magnitude of channel weights during training. %Other works such as
The authors in \cite{he2018soft} 
%and \cite{li2016pruning}
proposed pruning channels using thresholds by setting unimportant channels to zero. Network Slimming \cite{liu2017learning} used Lasso regularisation 
% on channel saliencies to induce sparsity and prune channels
with global thresholds. However, deep models pruned with structured sparsity methods lose their capabilities and connections permanently. As a result, dynamic channel pruning methods were devised that learn sparsity through a three-stage pipeline pretrain-prune-finetune or use pretrained models. The authors of \cite{gao2018dynamic} propose feature boosting and suppression (FBS) to dynamically amplify and suppress output channels computed by convolutional layers. \cite{tiwari2021chipnet} presents a deterministic pruning strategy using the continuous heaviside function and \textit{crispness loss} to identify a highly sparse subnetwork from an existing dense network.

\textbf{Conditional Channel Computation.} Regarding conditional computation at the channel level, the work proposed in \cite{lin2017runtime} generates decisions to skip computation for a subset of output channels. The channel gating network \cite{hua2019channel} finds regions among the features that contribute less to the classification result and skips computation on a subset of the input channels for these ineffective regions. 
% GaterNet \cite{chen2019you} proposes a separate gating network to learn channel-wise binary gates for the backbone network.
ConvAIG \cite{veit2018convolutional} introduced a network with a hard attention mechanism that adaptively selects specific layers of importance for each input image to assemble an inference graph by specifying a target rate for each layer. The authors of \cite{herrmann2020channel} also study conditional computation at the channel level and extend ConvAIG by learning target rates for each gate by specifying the target rate for the whole network. DGNet \cite{Li2021DynamicDG} proposed a dual gating mechanism by introducing sparsity along two separate dimensions, spatial and channel, in order to reduce model complexity at run time.
% Meng et al \cite{meng2020adafuse} introduced an approach to reuse history features along with dropping of unimportant channels to make the network capable of strong temporal modeling in case of video understanding.
For a more detailed background on sparsity, pruning and conditional computation, we recommend the review work presented in \cite{hoefler2021sparsity}.

% Pruning CNNs with Self Supevision : training from scratch, and not any pretrained model
While \cite{caron2020pruning} studied the behaviour of self-supervised models under standard pruning techniques, we investigate the effect of standard channel selection methods described in DGNet on self-supervised models. We also analyse whether networks trained under self-supervision with channel selection can preserve performance on downstream tasks. 
%Does channel selection guarantee transferability of learned representation over the downstream task? The need for channel selection during pretraining or during transfer? 
% Novel contributions of this paper - Placed at the end of Introduction section

\section{Method}

% In this section we provide details of augmenting self-supervised model with channel gating module.

\subsection{Self-supervised Module}

In this work we consider SimSiam \cite{chen2021exploring} as our self-supervised objective.  We use ResNet18 as a base encoder\footnote{across all experiments}, which takes two augmented views $\mathbf{x}_{1}$ and $\mathbf{x}_{2}$ from an anchor view $\mathbf{x}$ by applying stochastic augmentation from a set of augmentations $\mathcal{P}$. $\mathcal{P}$ comprises random resized crop, color jitter, random gray scale, Gaussian blur and random horizontal flip. These augmented views are processed through $f_{\boldsymbol{\theta}}$ to get a compact representation of $f_{\boldsymbol{\theta}}(\mathbf{x}_{1}), f_{\boldsymbol{\theta}}(\mathbf{x}_{2}) \in \mathbb{R}^{512}$. One view is further processed by a prediction MLP head (bottleneck architecture) $g_{\boldsymbol{\phi}}$ giving rise to an asymmetric architecture i.e.  $\mathbf{p}_{1} \triangleq g_{\boldsymbol{\phi}}(f_{\boldsymbol{\theta}}(\mathbf{x}_1)))$ and $\mathbf{z}_{2} \triangleq f_{\boldsymbol{\theta}}(\mathbf{x}_2)$. As a standard practise, a base encoder is augmented with a projection head MLP i.e., $f_{\boldsymbol{\theta}} = h \circ m, $ where $m$ and $h$ represents ResNet18 and projection layers respectively. The SimSiam  learning objective simplifies to a symmetric cosine similarity:
\begin{equation}
    \mathcal{L}_{\texttt{SSL}} = \frac{1}{2}\mathcal{D}(\mathbf{p_1}, \texttt{SG}(\mathbf{z_2})) + \frac{1}{2}\mathcal{D}(\mathbf{p_2}, \texttt{SG}(\mathbf{z_1})),
\end{equation}
% where,
%
% \begin{equation*}
%     \mathcal{D}(p_1, z_2) = -\frac{p_1}{||p_{1}||}. \frac{z_{2}}{||z_2||}.
% \end{equation*}
%
where $\mathcal{D}(\mathbf{a}, \mathbf{b}) = - \mathbf{a}^{T}\mathbf{b}$, with $\mathbf{a}$ \text{and}  $\mathbf{b}$ being $L_{2}$ normalised vectors\footnote{i.e. $\mathcal{D}(\mathbf{a}, \mathbf{b})$ is negative cosine similarity.}. \texttt{SG} stands for \texttt{Stop-Grad()}.

\subsection{Channel Selection via Gating}

% \begin{wrapfigure}{r}{0.25\textwidth}
% %  \vspace{-20pt}
%  \begin{center}
%     \includegraphics[width=0.2\textwidth, height=0.4\textwidth]{Screenshot from 2022-05-19 18-16-08.png}
% \end{center}
% \vspace{-20pt}
%  \caption{Introduction Banner.}
%  \vspace{-0pt}
% \end{wrapfigure}
\textbf{Preliminaries.}  Channel selection or conditional computation (data dependent gates) is often realised through a gating mechanism. A typical output for an input $\mathbf{x}$ from a convolutional (conv) layer $l$ is given by  $f_{l}(\mathbf{x}_{l-1})  \in \mathbb{R}^{C \times H \times W}$ where $f_{l}(\mathbf{x}_{l-1})$ consists of a convolution operation with kernel size $k$  followed with a batch normalization layer (BN) and relu  ($(\cdot)_{+}$) non-linearity with $\mathbf{x}_{l-1}$ being the output from the previous layer. The output from a gated convolutional network can be realised as: $\hat{f}_{l}(\mathbf{x}_{l-1}) = \pi_{l}(\mathbf{x}_{l-1})\cdot\text{BN}(\texttt{conv}_{l}(\mathbf{x}_{l-1}))_{+}$,
% To implement channel selection we follow \cite{Li2021DynamicDG}, where   $f_{l}(\mathbf{x}_{l-1})$ is augmented with a dynamic channel selection mechanism as shown below:
% \begin{equation*}
%     \hat{f}_{l}(\mathbf{x}_{l-1}) = \pi_{l}(\mathbf{x}_{l-1}).\text{BN}(\texttt{conv}_{l}(\mathbf{x}_{l-1}))_{+}
% \end{equation*}
where $\pi _{l}(\mathbf{x}_{l-1})\footnote{a vector of dimension equivalent to number of channels with ones and zeroes} \in \{0, 1\}^C $ is a gate dependent on input $\mathbf{x}_{l-1}$, which decides whether to keep (``on'')  or discard (``off'') a particular channel. This can be seen as a form of \textit{hard attention} (mask). This masking imposes a discrete structure over the network, making a computational graph for training and inference different. During training this structure is realised through stochastic gradient descent (SGD), while during inference it works as \textit{hard attention}. One of the main reasons for channel selection is to induce sparsity i.e. operate on a lower computational budget (less FLOPs) during inference. In this work we closely follow DGNet using ResNet18 as our base encoder.

\textbf{Channel Selection (Gating).} In order for gates\footnote{channel selection} to be effective, they need to estimate the importance of input features. This \textit{importance} is often referred to as relevance/saliencies (vectors) of the input feature map (along the channels) in the literature. This relevance is crucial in order for the network to avoid trivial solutions. A simpler way is to use SE block \cite{hu2018squeeze}, as was used in DGNet, to create a relevance vector. This usually requires getting a context vector $\mathbf{z} \in \mathbb{R}^{C}$ via global average pooling to accumulate spatial information. Finally, this context vector $\mathbf{z}$ is passed through a lightweight network to get channel attention $g_{l}(\mathbf{x}_{l-1})$, which can be summarized as:
\begin{align}
    g(\mathbf{x}_{l-1}) = \mathbf{W}_{1}\Big(\text{BN}\big(\mathbf{W}_{0} \mathbf{z}\big)\Big)_{+}, \quad \mathbf{W}_{1} \in \mathbb{R}^{C_{l} \times \frac{C_{l-1}}{r}},  \mathbf{W}_{0} \in \mathbb{R}^{\frac{C_{l}}{r} \times C_{l} },
\end{align}
where $r$ is a reduction ratio. For more details please refer to \cite{hu2018squeeze}. Finally, to achieve binary mask $\pi_{l}(\mathbf{x}_{l-1})$ we can use the channel attention $g_{l}(\mathbf{x}_{l-1})$ and set $\pi_{l}^{i}(\mathbf{x}_{l-1}) = 1$ if $g _{l}^{i}(\mathbf{x}_{l-1})\geq 0$ and $\pi_{l}^{i}(\mathbf{x}_{l-1}) = 0 $ otherwise. This discrete selection works during inference but it breaks the computational graph during training. To make the training possible, the Gumbel-SoftMax Trick \cite{jang2016categorical} is adopted. The Gumbel-Trick has been widely used as reparameterisation technique for the task of dynamic channel selection \cite{Li2021DynamicDG, herrmann2020channel, veit2018convolutional, meng2020adafuse}.
A gating block is introduced after the first convolution in \texttt{Basic Block} of Resnet18 following DGNet. Intuitively, the channel selection network could be interpreted as learning a policy whether to keep (compute) or discard (skip) a particular channel.

\subsection{Optimisation}{\label{opt}}
To remove unimportant channels and induce sparsity in the gating mask $\pi_{l}(\mathbf{x}_{l-1})$ we need to add an objective based on some budget $t_{d}$. To this end we use regularisation, a term used in DGNet as sparsity objective, which is a combination of sparsity and a bound regularisation term:
\begin{align*}{}
    \mathcal{L}_{G} = \lambda \underbrace{ \Bigg(\frac{\sum_{l=1}^{L}F^{R}_{l}}{\sum_{l=1}^{L}F_{l}^{O}} -t_{d}\Bigg)^{2}}_{\text{Sparsity}} + \gamma \mathcal{L}_{Bound}, 
\end{align*}
where $F_{l}^{R}$ is the average FLOPs over the batch along with FLOPs computation of the gating block, while $F_{l}^{O}$ is the original FLOPs without a gating module. Only the blocks with gating modules take part in FLOP computation as they are responsible for any sort of sparsity introduced in the network. The purpose of $L_{Bound}$ is to control early optimisation as detailed in DGNet.

\textbf{Final Objective.}  Overall training objective is defined as: $\mathcal{L} = \mathcal{L}_{\texttt{SSL}} + \mathcal{L}_{G}$, with $\lambda =5$ and $\gamma = 1$ across all the datasets and training regimes.  

% ---------------
 \begin{wraptable}{L}{0.5\textwidth}
 \vspace{-20pt}
% \begin{table*}[t]
\centering
  \caption{
  Performance comparison of SimSiam  with dynamic channel selection during inference. Evaluated with $k$-nearest neighbours ($k=1$) on the validation set of CIFAR-10, CIFAR-100 and ImageNet-100 across various target budgets $t_d$. \colorbox{LightCyan}{\textbf{*}} denotes baseline.} 
  \vspace{2mm}
  \label{tab2}
  \resizebox{8.5cm}{!}{
  \begin{tabular}{lcccc }
    \hline
    \hline
    % \multicolumn{1}{l}{}&\multicolumn{1}{c}{}&\multicolumn{1}{c}{}&\multicolumn{1}{c}{}\\
  
    \multicolumn{1}{l}{\textbf{Dataset}}&\multicolumn{1}{c}{\textbf{Budget} ($t_d$)}&\multicolumn{1}{c}{\textbf{Acc}$\%$}&\multicolumn{1}{c}{\textbf{FLOPs}} \\
    \hline
    % \multicolumn{10}{c}{}\\
    \hline
    \rowcolor{LightCyan}
    &*& 85.46\% & 7.03E8 \\
    &10\% & 76.72\% & 6.64E7 (90.55\%$\downarrow$) \\
    &20\% & 78.78\% & 1.25E8 (82.11\%$\downarrow$) \\
    &30\% & 80.82\% & 2.01E8 (71.41\%$\downarrow$) \\
    CIFAR-10&40\% & 81.35\% & 2.66E8 (62.18\%$\downarrow$) \\
    &50\% & 81.93\% & 3.29E8 (53.15\%$\downarrow$)\\
    &60\% & 82.96\% & 3.94E8 (43.89\%$\downarrow$)\\
    &70\% & \textbf{83.08}\% & 4.58E8 (34.76\%$\downarrow$) \\
    \hline
    \rowcolor{LightCyan}
    &*& 52.96\% & 7.03E8\\
    &10\% & 46.84\% & 6.88E7 (90.21\%$\downarrow$) \\
    &20\% & 49.50\% & 1.48E8 (78.88\%$\downarrow$) \\
    &30\% & 50.70\% & 2.03E8 (71.01\%$\downarrow$)\\
    CIFAR-100&40\% & 52.05\% & 2.65E8 (62.32\%$\downarrow$) \\
    &50\% & 52.54\% & 3.25E8 (53.67\%$\downarrow$) \\
    &60\% & 53.18\% & 3.95E8 (43.73\%$\downarrow$) \\
    &70\% & \textbf{53.50}\% & 4.66E8 (33.69\%$\downarrow$) \\
    \hline
    \rowcolor{LightCyan}
    &*& 64.34\% & 1.81E9\\
    &30\% & 56.08\% & 5.30E8(70.43\%$\downarrow$)\\
    ImageNet-100&40\% & 57.86\% & 7.13E8 (60.66\%$\downarrow$)\\
    &50\% & \textbf{60.38}\% &8.78E8 (51.55\%$\downarrow$)\\
    \hline 
  \bottomrule
\end{tabular}
}
\vspace{-12.5pt}
% \end{table*}
\end{wraptable}
% ---------------
\section{Experimental Setup}
% In this section we elaborate our experiment in order to validate the feasibility of dynamic channel modelling for self-supervised learning.

% However, we do not focus on pruning for faster computation i.e speedup during inference (in terms of MACS or FLOPS) instead we perform channel selection to investigate whether selecting less channels  still results in comparable performance.
\textbf{Implementation Details.} We closely follow the approach in DGNet for channel selection. For training, we use SimSiam as a self-supervised model with ResNet18 as a base encoder whose objective is modified as explained in section~\ref{opt}. We train the model with varying target densities $t_{d}$. The implementation of SimSiam is based on the solo-learn library \cite{JMLR:v23:21-1155}.  The base encoder is \textit{randomly initialised}\footnote{default initialisation in Pytorch} and is trained with SGD for 500 epochs (for a given target budget $t_d$) with a batch size of 256 on 2 Nvidia 2080Ti GPUs, with a warm-start of 10 epochs following a cosine decay with base learning rate of 0.01. Since we are using a very lightweight model as our gating network, there is no significant computational overhead during training. We report the inference speedup in terms of the hardware-independent theoretical metric of FLOPs and not wall-clock time as we are not using any hardware accelerators to utilise  sparsity during training. Code is made available \href{https://github.com/KrishnaTarun/SSL_DGC}{\textcolor{red}{here}}.

\textbf{Evaluation.} Training and evaluation is carried on train and validation data of CIFAR-10, CIFAR-100 and ImagNet-100 respectively.   For Cifar-10/100 we train for $t_{d} = \{0.1, 0.2, 0.3, 0.4, 0.5, 0.6, 0.7\}$ while for ImageNet-100 we restrict $t_d$ to only $\{ 0.3, 0.4, 0.5\}$ due to computational constraints. We use $k$-nearest neighbours as our evaluation metric evaluated with $k=1$. For \underline{\textit{baseline}} we train Simsiam with standard  objective without any channel selection for each of the datasets under consideration (* in Table \ref{tab2}).

\textbf{Results.} Our main findings based on the evaluation criteria validate our initial hypothesis that self-supervised models can learn highly redundant channel features. 

 Table~\ref{tab2} shows that in the case of CIFAR-10, by keeping only 70\% of the channels across the whole network, SimSiam achieves 83.08\% accuracy on the KNN task, which is a minor drop from the baseline performance of 85.46\% but at an ample reduction of 34.76\% in FLOPs. Furthermore, we also find that an enormous 90.55\% of FLOPs can be reduced by using only 10\% of the channels across the whole network causing a drop of only 8.74\%  in KNN accuracy. For CIFAR-100, we found that by restricting the channel usage to only 60\% over the whole network, SimSiam surpasses the baseline KNN accuracy of 52.95\% by 0.22\% reaching 53.18\%. Additionally, FLOP computation can be reduced by 90.21\% by keeping only 10\% of the channels, leading to a drop of only 6.12\%  in KNN accuracy.
On ImageNet-100, 50\% of the channel usage in the entire network results in 60.38\% KNN accuracy, which is 3.96\% less than the baseline. However, this decrease in accuracy is compensated by $\sim$51.55\% percent drop in FLOPs. Aside from this, we get a substantial 70.43\% drop in FLOPs by fixing channel utilization to only 30\% in the whole model. Therefore, channel selection can be thought of as a way to take advantage of the  trade-off between performance and computation depending the downstream task and individual use case. These results also show that SSL models trained with channel selection preserve the performance in downstream tasks.

Figure \ref{fig:1} shows the channel activation distribution for CIFAR-10, CIFAR-100 and ImageNet-100 datasets, revealing a deeper insight into the dataset specific behaviour of the channel selection network by visualising how many channels in each ResNet18 blocks are always off (skipped), always on (computed), or input dependent. We notice that significant number of channels are switched off and switched on all the time in all the three datasets and while others are input dependent. The channel distribution for CIFAR-10 and CIFAR-100 are very similar, which might be due to the fact that image statistics in both of these datasets are similar. %In order words, our networks learns a dataset specific policy for conditional channel computation and can be helpful to study the quality of datasets.

\section{Conclusion}
\begin{figure}
    \centering
   \includegraphics[width=1.0\textwidth]{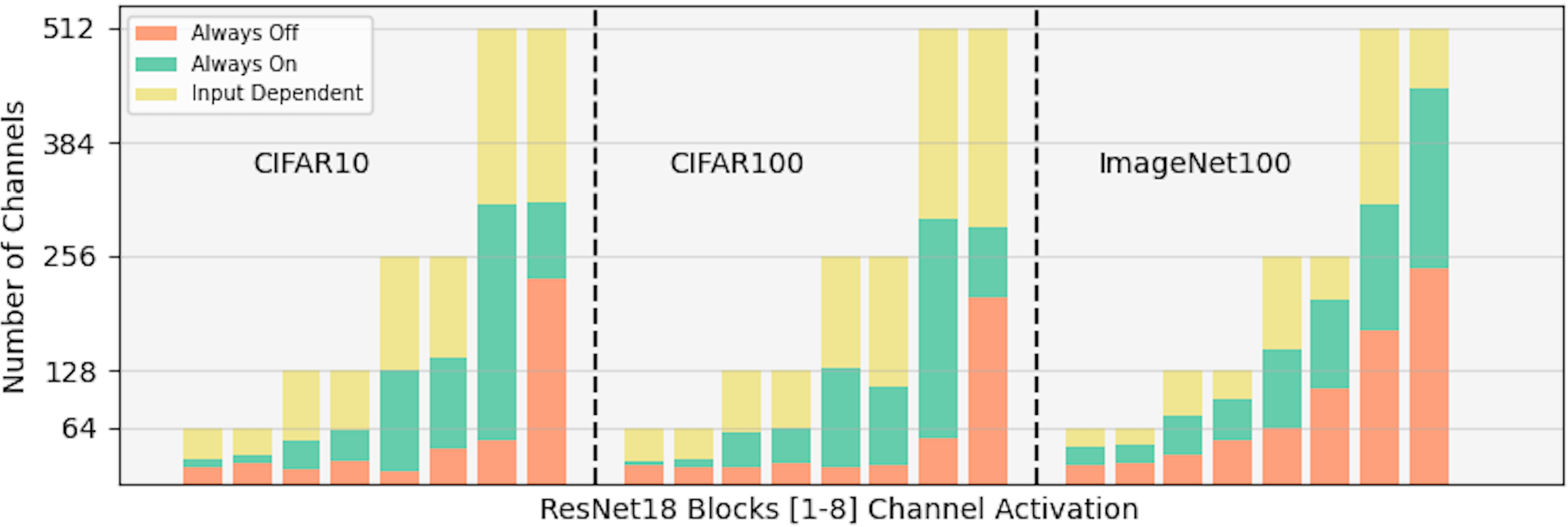}
    \caption{Channel distribution over validation set for $t_{d}=0.5$ on CIFAR-10, CIFAR-100, ImageNet-100}
    \label{fig:1}
     \vspace{-10pt}
\end{figure}

In this paper, we studied the behaviour of self-supervised learning when integrated with channel selection networks given a global target budget for computational cost. Our empirical results provided interesting insights about self-supervised learning when trained with channel selection. First, self-supervised models learn highly redundant channel features that can be discarded to reduce computational overhead (Figure \ref{fig:1}, Table \ref{tab2}). Second, we showed that channel selection modules can significantly reduce FLOP computation and make inference more efficient (Table \ref{tab2}). Third, our results also provide intuition that representations learnt by self-supervised networks with channel selection can also be transferred to downstream tasks.

% \begin{wrapfigure}{r}{0.6\textwidth}
%   \vspace{-30pt}
%   \begin{center}
%     \includegraphics[width=112mm, height=45mm,scale=0.5]{channel_plot.png}
% \end{center}
% \vspace{-30pt}
%   \caption{Channel Activation}
%   \label{fig1}
%   \vspace{-0pt}
% \end{wrapfigure}

There are, however, some limitations with our work. First, we still need to evaluate the tranferability of learned representations beyond classification to other downstream tasks such as object segmentation, detection and instance retrieval to name a few.  
% Secondly, we believe that consistency is an important factor for these self-supervised techniques, as almost every SSL method maximises the agreement between views (augmented views of the same scene/object i.e instance discrimination). Precisely this means two augmented views should have almost similar representations in the embedding space. We don't take this factor for channel selection into consideration by enforcing  some consistency aware constraints in the objective. These limitations will be addressed in our future work.
Second, the SSL training objective involves maximizing the agreement between augmented views of the same object or scenes (instance discrimination) and this forces them to have similar representations in the embedding space. In this work, we do not account for this by enforcing some consistency aware constraints for channel selection in the training objective. These limitations will be addressed in future work.

% we have not taken into account consistency between the two encoders in self-supervised models. 

%Note : The need for channel selection during pretraining or during transfer? Expand this

\section*{Acknowledgments}
This work has emanated from research supported by Science Foundation Ireland (SFI) under Grant Number SFI/12/RC/2289\_P2, co-funded by the European Regional Development Fund and Xperi FotoNation.

\bibliographystyle{apalike}

\bibliography{sample}

\end{document}